# ABCD: Trust enhanced Attention based Convolutional Autoencoder for Risk Assessment

SARALA.M.NAIDU*

Dept. of Innovation, Design and Technology, Mälardalen University, Västerås, Sweden

Power Grids, Grid Integeration, Hitachi Energy, Västerås, Sweden

Ning Xiong

Dept. of Innovation, Design and Technology, Mälardalen University, Västerås, Sweden

Anomaly detection in industrial systems is crucial for preventing equipment failures, ensuring risk identification, and maintaining overall system efficiency. Traditional monitoring methods often rely on fixed thresholds and empirical rules, which may not be sensitive enough to detect subtle changes in system health and predict impending failures. To address this limitation, this paper proposes, a novel Attention-based convolutional autoencoder (ABCD) for risk detection and map the risk value derive to the maintenance planning. ABCD learns the normal behavior of conductivity from historical data of a real-world industrial cooling system and reconstructs the input data, identifying anomalies that deviate from the expected patterns. The framework also employs calibration techniques to ensure the reliability of its predictions. Evaluation results demonstrate that with the attention mechanism in ABCD a 57.4% increase in performance and a reduction of false alarms by 9.37% is seen compared to without attention. The approach can effectively detect risks, the risk priority rank mapped to maintenance, providing valuable insights for cooling system designers and service personnel. Calibration error of 0.03% indicates that the model is well-calibrated and enhances model's trustworthiness, enabling informed decisions about maintenance strategies.

## 1 INTRODUCTION

### 1.1 Cooling System

HVDC technology has revolutionized long-distance electricity transmission, offering unmatched efficiency and reduced transmission losses. However, the power conversion process in HVDC systems generates substantial heat, posing a significant challenge to system reliability. This highlights the paramount importance of a robust cooling system in maintaining the integrity of HVDC stations. To effectively manage and monitor HVDC operations, Supervisory Control and Data Acquisition (SCADA) systems play a pivotal role. These systems continuously gather data on the station's operational status, providing a rich repository of information for data-driven risk identification and monitoring strategies. Within the HVDC station, various parameters are monitored based on specific functional areas and potential failure modes. The cooling system stands out as a cornerstone of system reliability, ensuring the efficient dissipation of heat and maintaining safe operating conditions for sensitive power electronics components. Among the parameters monitored in these systems, cooling liquid conductivity holds particular significance. Elevated conductivity signifies the presence of dissolved ions, potentially leading to performance degradation and increased risk of failures. Traditional SCADA monitoring methods for HVDC cooling systems often rely on fixed thresholds and empirical rules, which may not be sensitive enough to detect subtle changes in system health and predict impending failures. To address this limitation, an approach utilizing attention based convolutional autoencoders (CAEs) is proposed to detect risks. CAEs are a type of neural network that can learn the latent patterns in data and reconstruct the original data from a compressed representation. This ability makes them well-suited for identifying deviations from normal behavior, which can be indicative of potential system issues. Adding attention to the CAE would enrich to gain information on which feature to give importance to. To date, there is a scarcity of research specifically addressing anomaly detection in HVDC cooling systems.

* Corresponding Author. The work has been supported by the Swedish Foundation for Strategic Research (Project: ID20-0019)

The proposed approach represents a novel and promising method for enhancing HVDC system reliability. By leveraging the power of data-driven techniques, a deeper insight into system health gained to proactively address potential risks before they lead to failures.

## 1.2    Risk baseline

The parameter conductivity is measured using an analog meter that has a display range of 0.00 to 2.00 µS/cm and in SCADA the upper and lower control limits that are configurable for multiple alarm levels and trip levels. The SCADA system tries to maintain the cooling water conductivity as low as < 0.1µS/cm (at 25°C) [3]. Ion-exchange resin typically has an expected lifespan of several years. However, continuous exposure to high-conductivity makeup water can significantly shorten its lifespan. A rise in coolant conductivity during normal operation indicates the need to replace or regenerate the ion exchanger [2, 4]. According to preventive maintenance manuals [5], the ion exchanger should be replaced when the high conductivity alarm is triggered or at least every three years. Maintaining a regular replacement schedule ensures optimal system performance and effective impurity removal. To facilitate maintenance and upkeep, a scheduled planned shutdown for servicing is typically conducted annually. During this shutdown, a comprehensive system check is performed as per the Preventive Maintenance Program. Additionally, to ensure accurate conductivity readings, the sensor in direct contact with the coolant should be cleaned regularly using alcohol [5].

This risk data from the current practice serves as a reference baseline, which will be taken up in the results and discussion section for evaluation and validation with the output from the proposed data-driven model. The current state-of-the-art is to adopt the preventive maintenance schedules that follows the traditional recommendations, practices and timelines published from the OEM in the maintenance manuals. And the common trend evolved from the practice is to replace the ion exchange bottles during annual maintenance shutdown, to be on the safer side with a thought that "it is better to replace early than to be called to site for an investigation on high conductivity values". This is supported with the fact that a trip or stop of the key operations (power transmission) will cost a lot than to replace it upfront.

## 1.3    Learning representations

The SCADA data encompasses spatial correlations among different components of the cooling system, reflecting their interconnectedness. Temporal correlations also exist due to changes in operating conditions over time. However, the data primarily represents normal operation scenarios, making it challenging to obtain sufficient abnormal data. This imbalance and the lack of labels for the continuous operation data pose challenges in anomaly detection. Various approaches can be employed to identify anomalies, including one-class classification, novelty detection, out-of-distribution analysis, and reconstruction-based methods.

Reconstruction-based methods assume that the input parameters are correlated and utilize a machine learning model to embed the data into a lower-dimensional space, where normal and abnormal data exhibit distinctive characteristics. To understand the temporal dynamics of the latent space data, its temporal evolution needs to be computed. Nonlinear dimensionality reduction techniques, particularly neural networks, have gained popularity in this context. Autoencoders (AE) and its variant including the hybrid forms are very popular for different applications in the Energy domain, especially to tackle unsupervised learning problems, and eliminate complexity within datasets [6,7]. Deep autoencoders are used in anomaly detection in a semi-supervised way on data with no anomalies [13]. Due to its rich representation power with ability to learn characteristics the convolutions neural networks (CNN) have had high performance in multiple application areas not only in computer vision, but also in other domains [17]. While autoencoders learn abstract and high-level features, the CNN focus on low level features to captures the overall structure or content of the input data. The hybrid convolutional autoencoder (CAE) specifically have emerged as a powerful tool for anomaly detection due to their ability to learn and extract meaningful features from complex data in various applications [18,19]. Another aspect of the architecture is the attentions that has been extensively studied [14, 15] to focus on the important parts of the sequence and to improve the representations. [14] attention mechanism determines weights for each input word, contributing to a context vector that aids subsequent decoding steps. The application of attention-based CAE methods for feature learning, process monitoring, 3D variational data assimilations and other process is presented in [8, 9]. Attention-based AEs, with their ability to learn from changing data distributions, can adapt their models accordingly to maintain accurate monitoring and risk assessment even under dynamic conditions. Attention-based AEs, with their ability to capture both local and global features, can effectively extract these latent relationships and provide insights into the underlying dynamics of the system.

To address the mentioned challenges for the use-case, the proposed Trust enhanced Attention based Convolutional Autoencoder for Risk Assessment: ACBD framework is presented in this paper. The key contributions of work are summarized as follows:

1. A hybrid convolution autoencoder based neural network with attention is proposed to identify anomalies, by learning the comprehensive representations from the cooling system data.
2. The anomalies are then evaluated for potential risks and the risk priority rank or risk value are estimated.
3. With the estimated risk value, the link to its influence on the maintenance of the components is derived.
4. The calibration error is derived to increase the confidence for the model's prediction.
5. Experiment is conducted on a real-HVDC station cooling system data revealing satisfactory performance and application prospects of the proposed method.
6. The performance is compared to the convolution autoencoder without attention, to showcase the importance and the value add with attention layer.

To the best of the author's knowledge, attention-based convolution autoencoder for risk identification is for the first time being used in the cooling system field. Thus, this paper will be first novel baseline on the concept application. The rest of the paper is organized as follows: Section 2 describes the methods for risk detection, mapping to maintenance, followed by the evaluation of the model's prediction with calibration error, data and the training parameters followed by the discussion of the results and conclusion in section 2 and 4 respectively.

## 2  PROPOSED METHOD

In multi-sensor time series data, anomalies can occur at irregular intervals, and the data driven models must learn to identify these patterns without prior knowledge. Self-supervised learning techniques can be used to address this challenge by training the model to reconstruct the data from corrupted versions. The samples that deviate significantly from the expected reconstruction patterns are flagged as potential anomalies.

## 2.1 Network structure of ABCD:

By leveraging convolutional layers, a convolutional bottleneck, CAEs effectively capture intricate patterns and relationships within complex systems. In the context of the cooling system, CAE offers a unique advantage in transforming the high-dimensionality of the systems into a lower-dimensional latent space, enabling deeper insights into their behavior. The encoder portion of a CAE maps the original data onto this latent space, while the decoder reconstructs the original data from the latent representations. Due to the capability of handling spatial, spectral, and temporal information, the performance of CNN based autoencoders have been exceptional. To effectively capture spatial relationships in the datasets, CAEs employ one-dimensional convolutional layers. These layers utilize 1D kernels of a specified size to slide across the input data, extracting relevant features and preserving local correlations. The output of the convolutional layer is a feature map, which serves as the input to subsequent layers in the CAE architecture. Each datapoint x∈X is mapped to a corresponding datapoint y∈Y through a convolution operation. However, it tends to give equal importance to all features irrespective of the contributions to final task (be it classification or anomaly detection). The attention mechanism enhances the corresponding representational values by giving more weights to the important parts and decreasing weights for the unnecessary ones, i.e., the attention layer helps to focus on the important parts of the inputs, leading to a better transformation of features into the latent space. Thus, the overall architecture shown in Fig 1, integrates the following:

1. Convolutional kernel and autoencoder to deals with the variables to extract the intrinsic features and the spatiotemporal information.
2. An attention mechanism embedded between the encoder and decoder, to select the critical information and transmit it to the next layer.

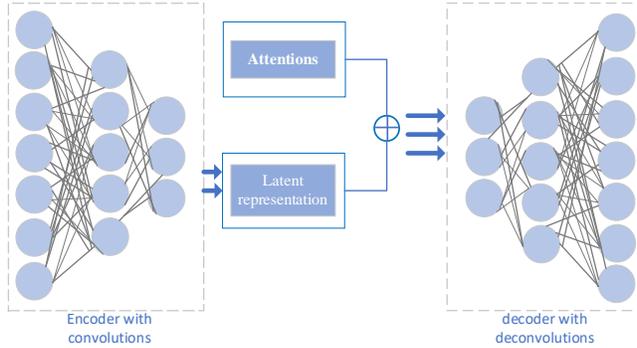
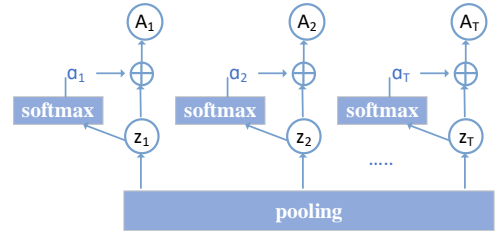

Fig 1: Network architecture of ABCD                     Fig 2: Attention computation process

The encoder consists of several layers of convolutions and pooling layers to encode the input data. The decode is a replica of the encoder consists of deconvolutions and up-sampling layers to decode the encoded features. The attention layer embedded between the encoder and the decoder performs the selection of the features. <u>Convolutional layers</u>: If $w_i$ is the weight parameter of the ith convolution kernel, where i is the number of channels, ⊖ is the convolutional operation and activation function f(x) given by Eq. (2), then the output of the convolutional layers that extracts the features, is given by Eq. (1).

$$C_i = f(\Sigma x ⊖ w + b_i) \qquad (1)$$
$$f(x) = ReLU(x) = max(0, x) \qquad (2)$$

<u>Pooling layer</u>: it is used to reduce the dimension to decrease the computational complexity. The output of the pooling with W, as the width of the pooling and S as stride, is given by Eq. (3).

$$P_i(m) = max \{C_i(mW, (m+1)W\} , 0 \leq m \leq \frac{L}{S} \qquad (3)$$

<u>Attentions</u>: used to pay attention to the critical features during training and decrease the weights of the unnecessary features. It dynamically adjusts the weights of output in the encoding phase and select critical features to reconstruct the input. This computation process is shown in Fig 12 and is given by Eq. (4).

$$c_i = f(\Sigma \times \ominus \omega_i + b_i) \qquad (4)$$

$$A = \sum_i \alpha_i \cdot z_i \qquad (5)$$

$z_i$ is the encoder extracted features, weight of the calculated attention is given by $\alpha\_i$ and A being the final output vector after attention.

<u>Deconvolutional layer</u>: In decoder the size of features is increased by deconvolutional layer as given by Eq. (6).

$$D_i = Re\,Lu(\sum x \oplus \bar{w}_i + C_i) \qquad (5)$$

Where $\oplus$ is the deconvolutional operation and $\bar{w}_i$ is the weight parameter of the ith deconvolutional kernel.

<u>Loss Function</u>: After the model has processed the input data and generated an output prediction, it compares the prediction to the actual value (label) using the mean absolute error (MAE). This comparison quantifies the error between the predicted and actual values, providing feedback to the model about its performance. The MAE serves as a loss function, guiding the model's optimization process through backpropagation. Backpropagation updates the model's parameters in a direction that minimizes the MAE, effectively steering the model towards better predictions in subsequent iterations. MAE cost function is calculated as in Eq. (6).

$$mae = \frac{1}{n}\sum_{i=1}^{n}|y_i - x_i| \qquad (6)$$

## 2.2    Anomaly detection:

Fig 3, represents the overall proposed ABCD framework. The anomaly detector is initially trained on normal data sequences, treating the task as a regression problem, where the model predicts continuous values. With each iteration, the model learns to capture the activation features and saliency features of normal data, leading to a decrease in prediction errors. After the model effectively internalizes the encoded latent space representation and reconstruction for normal data, mae is analyzed to establish an efficient threshold to distinguish between normal and anomalous data. This threshold is set to one standard deviation above the mean loss of the trained normal data samples, differentiating between normal and anomalous patterns.

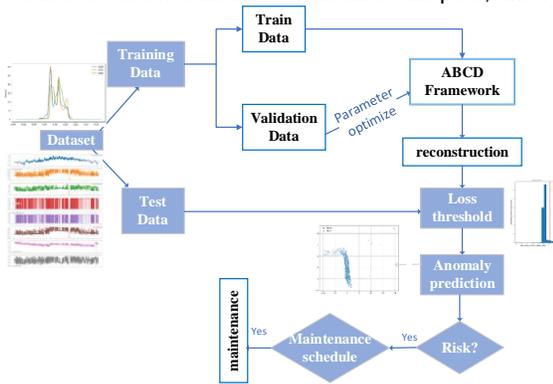

Fig 3: Overall architecture of ACBD framework

Then the threshold dictates if the data is normal or abnormal by comparing its reconstruction error, i.e., if the reconstruction error of the test data is higher than the threshold, it is labelled as anomalous.

## 2.3 Anomaly to risks map

Failure Mode Effect Analysis (FMEA)/Failure Mode Effect and Criticality (FMECA) is a systematic approach to establish how the function of an item fails, the consequences of failure and the detection methods to mitigate the effects of the failure. The FMECA process is derived according to IEC60812: 2018 [16]. In practice, three types of analysis are performed viz. severity of failure(S), probability of failure(P) and detection probability(D). To each anomaly detected by the model and confirmed by the domain expert, against each of the 3 analysis factors, a score in range of 1 to 5 is assigned with 1 being lowest and 5 being the highest probability/severity/detection, as given in Table 1. Equation (7) represent the risk value, or the risk priority rank (RPR) calculated based on the values of S, P and D.

$$\text{RPR} = S x P x D \tag{7}$$

RPR ranks the severity of the failure mode. Say, for example if the RPR <= 25, the domain knowledge would indicate risk is low, if 26 < RNP <= 50 risk is medium and if RPN >65 the risk is high. These values are adapted to domain process. The impact of failure with respect to the anomaly detected can be derived, with is step.

Table 1: Score description

| Description of probability | Very low | Low | Moderate | High | Very high |
|---|---|---|---|---|---|
| *score* | 1 | 2 | 3 | 4 | 5 |

## 2.4 Measurement of calibration

Calibration is a useful tool for making informed decisions about whether to accept or reject a model's prediction. This helps increase the confidence in the model's predictions. Among multiple ways of using the calibration like thresholding, confidence intervals, the calibration metric along with reliability diagram helps visualize the relationship between predicted and observed probabilities. If the points on the reliability diagram fall close to the diagonal line, then the model is well-calibrated. In simpler terms, for every input x and predicted y, the model's predicted probability p' should exactly match the actual probability of observing y, given the input x. This condition ensures that the model's predictions align with the underlying true probabilities, making its assessments more reliable and trustworthy. The model is considered well-calibrated if p' perfectly represents the true probability, represented mathematically by Eq. (8).

$$\text{For all values of x and y, } p'(x,y) = P(Y=y|X=x) \tag{8}$$

Any difference between the right side and the left side for a given p´, is called the calibration error. To effectively assess the calibration error of a model, Expected Calibration Error (ECE) divides the probability range into a set of intervals, or bins, and assigns each predicted probability to the bin it falls within. The calibration error for each bin is calculated as the difference between the accuracy (fraction of correct predictions) and the confidence (mean probability) within that bin. ECE computes the weighted average of the error across bins as in Eq.(9).

$$ECE = \sum_{b=1}^{B} \frac{n_b}{1^p} |acc(b) - conf(b)| \tag{9}$$

$n_b$ is number of predictions in bin b on N datapoints, accuracy and confidence of the bin b are acc(b) and conf(b).

## 2.5 Datasets

Data from the HVDC station's cooling subsystem was collected from 2020 to 2023 and preprocessed to ensure uniformity. The 2021 and 2022 data were used for training, while data from 2020 served as unseen test data. Standardization was applied to transform each feature to have a mean of 0 and a standard deviation of 1.

## 2.6 Training and parameters

80-20% strategy is applied to split data into train and test sets. The configuration of parameters for model training is as given in Table 2.

Table 2: Training parameters

| Parameters | Value range used for best results |
|---|---|
| TimeSteps | 24 hr |
| Batch size | 16 to 64 |

| Parameters | Value range used for best results |
|---|---|
| Learning rate | 0.01 to 0.0001 |
| Epochs | 100 to 500 with callback to monitor the loss, with patience=5, on a validation split of 10% |
| Optimizer | Adam |
| Activiation function | Relu, |
| No. of layers | CNN = 2 Layers, Encoder = 2 layers followed by bottle neck, Decoder = replica of encoder |
| Data | Hourly sampled data of 2 years (2021 and 2022) used for training and new year 2020 used for testing |

## 3 EXPERIMENT RESULTS AND DISCUSSION

### 3.1 Anomaly diagnosis

From random monthly data taken from year 2020 for testing, two example cases are taken for discussion here.

<u>Case-1</u>: The anomaly is identified starting on the 9[th] Dec-2020 at a time of 10:01:59, and followed by a sequence of 9 anomalies marked by the model as in Fig 4[A], with zooming in shown in [B] on the day of the anomaly sequence showing the multiple points of anomaly. The Fig 4[C] is the tabulated data identified by model as anomalous, with the timestamps for each. The duration of the anomaly sequence in this case is for 8250 milliseconds starting at a timestamp of 10:01:59.671 and ending at 10:02:07.921.

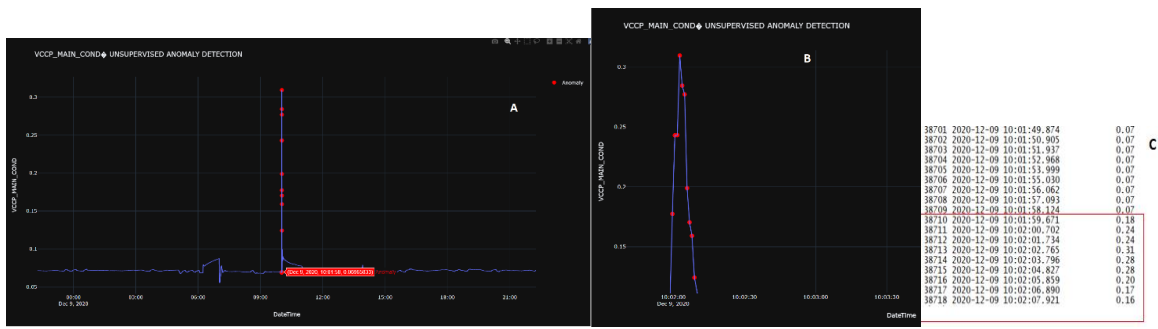

Fig 4: Case 1: Anomaly on 2020-Dec-09 10:01:59 accepted. [A] is overall anomaly, [B] is zoom in of the sequence of the anomalies, [C] is the tabulated anomaly detected data.

<u>Case-2</u>: Another random test case for a data of September 2020, with anomalies marked on the 24th and 26th of sept-2020 by the model. The Fig 5[A], encircled in yellow is a set of 12 anomaly points lasting for 14 seconds showing the quick and continuous increase in conductivity values almost every second as identified by the model, while the anomaly [B] encircled in green is a set of 3 points with low abnormal values that are negative and lasting for 3 seconds. In both cases, the plots show the normal data points, pre and post the anomaly occurrence that are correctly identified as normal.

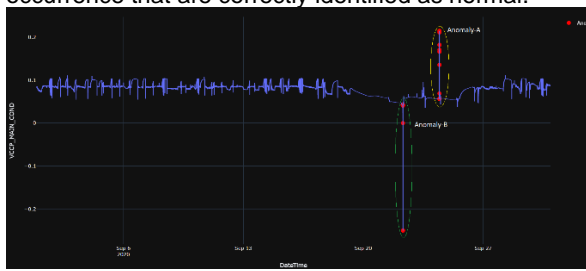

Fig 5: Case 2: Anomalies on the 24th and 26th of sept-2020

### 3.2 Performance and comparison

Mae results for comparing performance with and without attentions is recorded in Table 3, on same test data and hyperparameters.

Table 3: Training parameters

| Model | MSE score |
|---|---|
| CAE with NO attention | 0.0094 |
| CAE with attention | 0.00402 |

Thus, with the ACBD framework (the attention-based CAE) there is an increase in mae metric performance by 57.44% with the self-attention layer in the encoding phase as compared to the model without the self-attention layer. Also, there is a decrease in the number of false alarms by 9.375%, as the number of anomalies reduced from 32 in CAE (without attention) to 29 (with ABCD), on the test samples. Hence, on the given data and context, the proposed ABCD model has improved performance.

### 3.3    Risk map on confirmed anomalies.

Now the analysis of the cause in the case-1 anomaly on 9th Dec-2020, in Fig 4[A] and [B] it is expected (as per domain) that the cause for increase in conductivity should be increase in supply temperature. However, the supply temperature is in normal range. Now for case-2 anomalies in Fig 5[A], the increase in the conductivity for 14 seconds, is due to the increase in the supply temperature closer to 40°C and back to normal values with the decrease in temperature. And the case of anomaly marked as [B] in Fig 5 for 3 seconds, is due to disturbances in the sensor measurements, as it corresponds to the similar negative values in the supply temperature data. Thus, from domain point of view it is a false alarm, but a real anomaly from data perspective. There is a clear cause of increase in conductivity values, with increase in supply temperature. And this cause is for a short duration and is not seen to re-occur consistently in the time span. As detailed in the "unpublished" work in [12] the explanations on model's output can be leveraged using the explainable AI (XAI). The event of increase is rare as seen in the 3 years data span, and anomaly reason for case-1 is unknown as there is no corresponding increase in supply temperature as in case-2. In the maintenance documents [3,5] it is not stated to what duration the increase in conductivity is allowable without risk. Result of this experiment can be a use-case to bring out this information during the system design and influence the recommendation from the original equipment manufacturers (OEM) of the cooling system components like ion-exchange vessel, resin etc. Table 4 shows risk value estimated based on severity, probability of failure and detection probability from Eq. (7).

Table 4: Risk Values

| Description | Value description | value |
|---|---|---|
| No. of occurrence of High alarm level -1 & 2[a] | In span of 3.5 yrs (2020 to  jun 2023)<br>{*for a duration of 1.03 sec from Fig 5[C]} | 1 * |
| No. of occurrence of High alarm level -3[a] | In span of 3.5 yrs (2020 to  jun 2023) | 0 |
| No. of occurrence of low alarm (value close to 0 ) | In span of 3.5 yrs (2020 to  jun 2023) | 0 |
| No. of occurrence of abnormal sensor readings | In span of 3.5 yrs (2020 to  jun 2023) | 3 |
| S=severity of failure | very high severity due to its impact of trip | 5 |
| P= probability of failure | high probability, as failure can occur  weekly | 4 |
| D=detection probability | Very high detection probability if ontinuously monitored. OR<br>High detetction probability as it can be identified quickly as it occurs | 2<br>1 |
| RPR | SxPxD | 5x4x2=40 OR<br>5x4x1=20 |

Thus, the risk value is moderate if the probability of failure is continuously monitored, and the risk is low if it can be identified quickly once it occurs. This analysis aligns with the analysis and confirmation from SME, as there was no trip related to the high alarm level 3 in the time span of 3 years. This could be inferred that the ion-exchange process is performing very well in treatment process.

### 3.4     Measuring calibration

Calibration is used to accept or reject a model's prediction, as it helps to ensure that the model's predictions are reliable and trustworthy. A well-calibrated model has predicted probabilities that closely match the actual outcomes. Fig 6 summarizes how the predicted probabilities aligns with the observed frequencies in the experiment. For example, at a predicted probability of 0.5, the observed frequency is 0.45. This means that the model is predicting that the event will occur 50% of the time, but it is occurring 45% of the time.

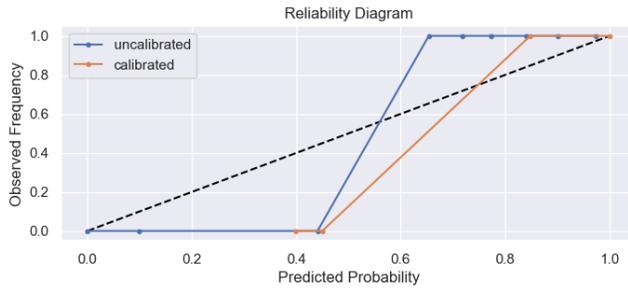

Fig 6: Reliability diagram (calibration curve)

Here, the calibration error is 0.03%, means that the model's predicted probabilities are very close to the actual observed frequencies. In general, a calibration error of less than 0.01 is good. Calibration error is just one measure of a model's performance and is typically calculated for a specific bin size. Thus, a calibration error of 0.0003 is very good value and indicates that the model is well-calibrated. Other adaptive calibration errors in [11] can also apply here.

### 3.5     Risk Influence on maintenance.

With reference to the background and baseline in section I.A.1), an increasing trend of the coolant conductivity during regular operation is a sign that the ion exchanger bottle needs to be replaced or that the exhausted resin should be changed. To avoid an outage or trip, the recommendation is to replace the ion-exchange resin when the high conductivity alarm is triggered or at least every three years. However, the current practice is to replace it during the planned annual maintenance every year. Now with the analysis of the results from the data driven approach using the ACBD framework, the risk value, the reliability score of the model's prediction and the duration of risk are very useful information to enrich the cooling system designers and the service personnel for decision making. With the status of the current conditions of the conductivity increase trend and the risk values, the resin replacement can be delayed from the current schedule. It can be replaced safely only based on the current condition of the coolant conductivity. This also helps save cost on the material and effort of the service personnel for handling ion-exchange.

## 4   CONCLUSION

This paper proposes a novel attention-based convolutional autoencoder framework, ABCD, that effectively detects anomalies in cooling liquid conductivity and identifies potential risks in HVDC systems. The framework's ability to learn the normal behavior of cooling liquid conductivity from historical data enables it to accurately detect deviations from the expected patterns, which can serve as early indicators of impending failures.

Moreover, ABCD can effectively map these anomalies to maintenance schedules, providing cooling system designers and service personnel with valuable insights into the timing of necessary interventions. To ensure the reliability and trustworthiness of the model's predictions, a calibration technique to quantify the match between predicted probabilities and actual observations is employed. Well-calibrated models produce predictions that align closely with real-world outcomes, enhancing confidence in their assessments.

However, decision-making processes must carefully consider the costs and effort associated with replacing ion exchange resin against the risk insights derived from data-driven condition assessments. A thorough evaluation of these factors is essential to make informed decisions that optimize maintenance strategies and minimize overall system downtime. Future research directions include incorporating continuous and online learning mechanisms into the ABCD framework. This would allow the model to continuously adapt and update its knowledge as new data becomes available, further enhancing its effectiveness in detecting anomalies and predicting risks.

## ACKNOWLEDGMENTS

We acknowledge the support by the Swedish Foundation for Strategic Research (Project: ID20-0019).

**1. Please help us understand your paper better by completing below form, and it will not be published.**

| First Author | Position : Ms. Sarala M Naidu |
| --- | --- |

|  | **Research Field:** Self Supervised Learning methods for Predictive maintenance in Power systems |
|  | **Homepage URL**: https://www.es.mdu.se/staff/4635-Sarala_Mohan |
|  |  |
| **Second Author** | **Position: Prof. Ning Xiong** |
|  | **Research Field: Learning and Optimization** |
|  | **Homepage URL:** https://www.es.mdu.se/staff/82-Ning_Xiong |
|  |  |
| **Add more rows if necessary!** | |